# Dental CLAIRES: Contrastive LAnguage Image REtrieval Search for Dental Research

**Tanjida Kabir, MS[1], Luyao Chen[1], Muhammad F Walji, MS, PhD[2], Luca Giancardo, PhD[1], Xiaoqian Jiang, PhD[1], Shayan Shams, PhD[1,3]**

[1]University of Texas Health Science Center at Houston, School of Biomedical Informatics, Houston, Texas, USA; [2]Department of Diagnostic and Biomedical Sciences, The University of Texas Health Science Center at Houston, School of Dentistry, Houston, Texas, USA; [3]Department of Applied Data Science, San Jose State University, San Jose, California, USA

**Abstract**

*Learning about diagnostic features and related clinical information from dental radiographs is important for dental research. However, the lack of expert-annotated data and convenient search tools poses challenges. Our primary objective is to design a search tool that uses a user's query for oral-related research. The proposed framework, Contrastive LAnguage Image REtrieval Search for dental research, Dental CLAIRES, utilizes periapical radiographs and associated clinical details such as periodontal diagnosis, demographic information to retrieve the best-matched images based on the text query. We applied a contrastive representation learning method to find images described by the user's text by maximizing the similarity score of positive pairs (true pairs) and minimizing the score of negative pairs (random pairs). Our model achieved a hit@3 ratio of 96% and a Mean Reciprocal Rank (MRR) of 0.82. We also designed a graphical user interface that allows researchers to verify the model's performance with interactions.*

## 1. Introduction

More than 1.4 billion dental radiographs are taken annually in the United States for regular oral health exams and to diagnose dental abnormalities[1,2]. These radiographs are often associated with one or more diagnoses and patient information such as demographics, medical, treatment history, and other necessary findings[3]. Despite their high research value, existing data repositories lack an easy-to-use interface for researchers to search radiographs based on complex and heterogeneous medical information[3,4]. Natural language processing is highly convenient and flexible to specify complex conditions but achieving a highly accurate medical image retrieval task is nontrivial. Furthermore, simple solutions to compare only text queries with information associated with radiographs might not yield the best performance because the vital connection between the radiograph contents and their association with clinical information is often not fully utilized.

The recent development of deep learning technologies makes it possible to explore the latent connections between image and text. Several state-of-the-art methods were designed for content-based image retrieval using natural image datasets. For example, Contrastive Language-Image Pre-Training (CLIP) is trained on various image-text pairs (1.28 million training examples) to predict the most relevant text on a given image[5]. CLIP has achieved remarkable success thanks to the abundant image-text pairs in the general domain. Quattoni et al. predicted words associated with images utilizing the manifold learning process on natural image data[6]. Joulin et al. showed that convolutional neural network (CNN) models could speculate words for captions by learning useful image features[7]. Dong et al. developed a deep neural network to predict visual features from the text for images and videos[8]. Most of these models focus on visual question answering (VQA) to answer natural language questions given a particular image, which is a relevant but different problem to our task (i.e., retrieving relevant images with natural language queries).

Few studies have been conducted to generate automated search systems in the medical domain. In particular, Zhang et al. showed a deep learning-based framework to learn visual representations by utilizing image-text pairs for chest and musculoskeletal radiographs[9]. Songhua developed a search engine for biomedical images and associated papers based on the user's query[10]. Eslami developed a fine-tuned version of CLIP, PubMedCLIP, for the medical image-text pair ROCO dataset and gained 3% accuracy over the state-of-the-art model[11]. This study showed the necessity for developing domain-specific image-text pair search models. However, to the best of our knowledge, no studies have been conducted to learn visual representation from the raw text for dental image retrieval. Therefore, this study aims to design an automated search engine for dental radiographs utilizing the current knowledge base for secondary research and patient dental care.

We propose a framework called **C**ontrastive **LA**nguage **I**mage **RE**trieval **S**earch for dental research (Dental CLAIRES) to predict the periodontal images described by the user's texts. In this study, we have jointly utilized periapical radiographs and related metadata describing the periodontal diagnosis, position, and demographic information obtained from the electronic health record (EHR). These textual data help to learn necessary visual features from images. Generally, periapical radiographs are treated as a 'gold standard' and primary examination method for dental abnormalities[12], especially periodontitis, since they can capture accurate and detailed anatomical structures and bony defects and are available in almost all dental clinics. To the best of our knowledge, it is the first work to combine dental radiographs with natural language to predict the most relevant images represented by text snippets. It can reduce the healthcare burden by combining imaging and EHR data.

The contrastive representation learning[13] method is utilized by maximizing the score of true image-text pairs and minimizing the randomly generated pairs. Furthermore, we performed two ablation studies-utilizing texts only and images only for searching and compared their performances to illustrate the effectiveness and robustness of the proposed model. In addition, we have designed a graphical user interface (GUI) for researchers to provide feedback on the usefulness and usability of the framework. The typical outline of the proposed Dental CLAIRES model is that users can provide keywords, such as periodontal stage and region, to retrieve the most relevant images.

## 2. Methods

### 2.1 Overview of the proposed system

This study was conducted following the guidelines of the World Medical Association's Declaration of Helsinki and the study checklist for artificial intelligence in dental research[14] and approved by the University of Texas Health Science Center at Houston Committee for the Protection of Human Subjects (HSC-DB-20-1340).

Figure 1a illustrates the high-level architecture of the training process of the proposed framework. Here the model was trained on a set of image-text pairs to predict the most relevant image given by the texts. First, a set of texts A and a set of images B were considered where $(A_i, B_i)$ and $(A_i, B_j)$ were the true positive pairs and negative random pairs, respectively. Next, the images/text went through the images/text encoders and projection functions to generate the embedded representations of the images/text. Then, the similarity between all the image-text embedding pairs (true and random) was calculated using the cosine similarity function. Next, the proposed model is trained to generate the best-embedded representations by maximizing the agreement between the true image-text pairs and minimizing the agreement between randomly coupled pairs. Finally, the trained model returns the best-matched images narrated by the user's query (Figure 1b).

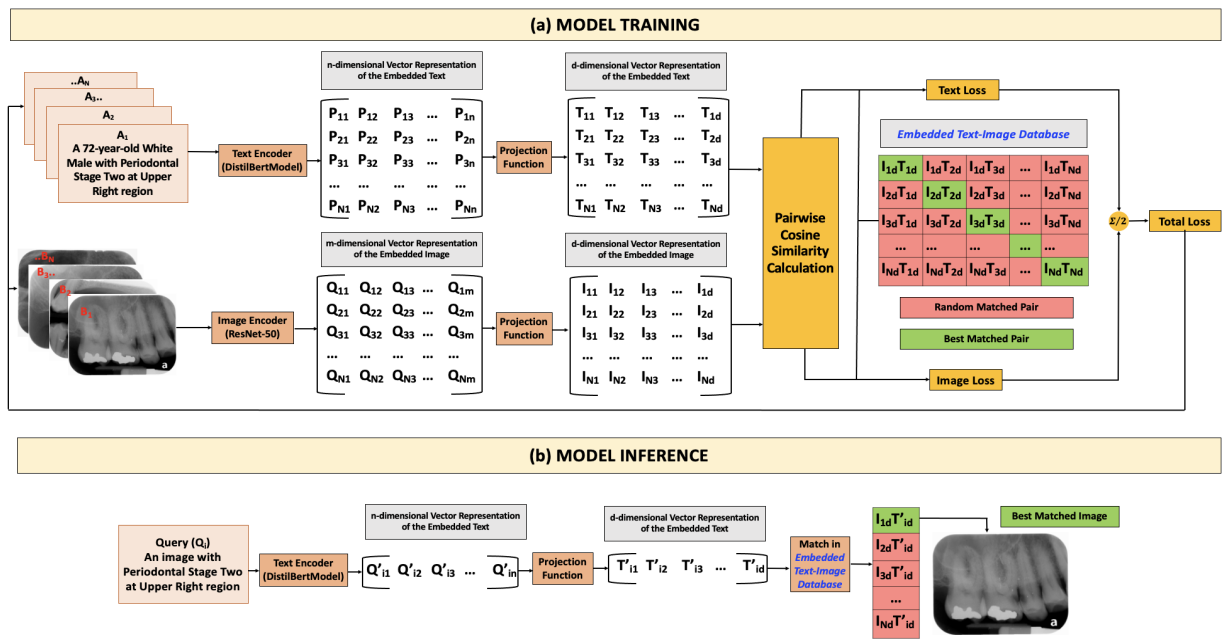

**Figure 1:** (a) The proposed Dental CLAIRES model's training steps using periapical radiographs and related clinical information. (b) Model inference steps to find the best-matched image described by the texts.

## 2.2 Dataset Description

687 periapical images of 45 randomly selected periodontitis patients from the UTHealth School of Dentistry were used in this study. The selection criteria were adult subjects ($\geq 18$ *years old*) diagnosed with periodontitis. Three independent examiners (a periodontist professor, a clinical periodontist, and a periodontal resident) annotated the images for periodontal staging and the radiographs' positions, such as upper molar right, upper molar left, lower molar left, lower molar right, lower anterior, and upper anterior. Each examiner did the annotation independently. If there was a conflict among annotators for periodontal stage assignment, then majority voting was utilized for final staging. Assigning stages using the radiographic bone loss (RBL) percentage is based on the 2018 periodontitis classification[15].

- o Stage 1: RBL < 15% (in the coronal third of the root)
- o Stage 2: 15% $\leq$ RBL $\leq$ 33% (in the coronal third of the root)
- o Stage 3: RBL > 33% (extending to the middle third of root and beyond)

Usually, stage 1 is considered as control group and stage 2 and 3 are considered as treatment group for periodontal evaluation. In addition, demographic information such as age, gender, and ethnicity were collected from the EHR data. Then we combined that demographic information with periodontal stages and regions to obtain a descriptive text for each image. Figure 2A represents samples of image-text pairs for different periodontal stages and their corresponding demographic information extracted from the EHR database. Figure 2B provides the dataset distribution used in this study.

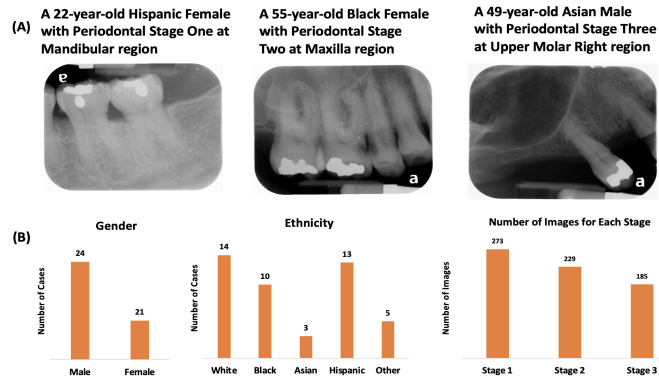

**Figure 2:** (A) Sample image-text pairs for different periodontal diagnoses and (B) Dataset distribution in gender, ethnicity, and different periodontal stages for the selected cohorts. *The captions are synthetic for patients' privacy

## 2.3 Image Augmentation and Text Variation

Image augmentation processes such as rotation and contrast variation (Figure 3) were utilized to make the dataset more generalizable. In addition, multiple variations of single texts, such as synonyms of different facial areas, including and excluding demographic information, were generated (Table 1). The synonyms were replaced using python script. For example, the original EHR text was: "A 72-year-old White female with Periodontal Stage Three in the Maxilla region". The maxilla is a pair of irregularly shaped bones forming the upper jaw (https://www.ncbi.nlm.nih.gov/mesh/68008437). We generate new text, "A 72-year-old White female with Periodontal Stage Three in the Upper Jaw region," by replacing the maxilla with the upper jaw. The synonyms were extracted from the PubMed database and general dental literature. The motivation for introducing synonyms for texts is to make the framework robust and more generalizable. Using image augmentation and text variation, we generated 30 new image-text pairs (5 *images* $\times$ 6 *captions*) for each original image-text pair.

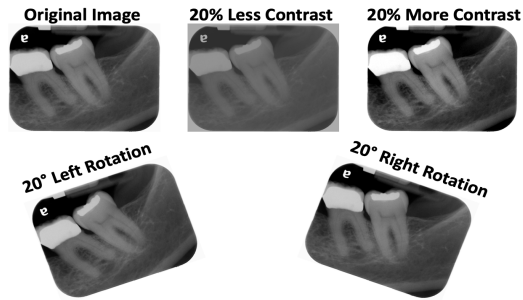

**Figure 3:** Image augmentation process was adopted to make the dataset more generalizable.

**Table 1:** The text augmentation (multiple variations of the texts) for a single image.

| Images | Describing Texts |
|---|---|
| 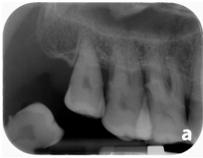 | **Caption 1**: A 72-year-old White female with Periodontal Stage Three in the Maxilla region.<br>**Caption 2:** A 72-year-old White female with Periodontal Stage Three in the Upper Jaw region.<br>**Caption 3:** A 72-year-old White female with Periodontal Stage Three in the Upper Molar Right region.<br>**Caption 4:** A White female with Periodontal Stage Three in the Upper Molar Right region.<br>**Caption 5:** An Image with Periodontal Stage Three in the Upper Molar Right.<br>**Caption 6:** An Image with Periodontal Stage Three. |
| 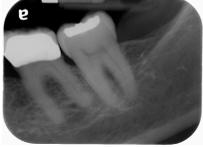 | **Caption 1**: A 29-year-old Black Male with Periodontal Stage Two in the Mandible region.<br>**Caption 2:** A 29-year-old Black Male with Periodontal Stage Two in the Lower Jaw region.<br>**Caption 3:** A 29-year-old Black Male with Periodontal Stage Two in the Lower Molar Left region.<br>**Caption 4:** A Black Male with Periodontal Stage Two in the Mandible region.<br>**Caption 5:** An image with Periodontal Stage Two in the Mandible region.<br>**Caption 6:** An image with Periodontal Stage Two. |

## 2.4 Model Description

We utilized the idea of the original CLIP[5] model in our proposed framework, where we utilized embedded image-text pairs to learn necessary features from images and associated texts. First, the encoder and projection function were applied to generate the embedded image-text pairs. Then we calculated the cosine similarity among all the image-text pairs to generate a text-image embedding database for future references (Figure 1a).

**Image and Text Encoder Models:** The DistilBERT[16] model was used for text encoding and the ResNet-50[17] model was used for image encoding. DistilBERT[16], a lighter version of BERT[18], is 60% faster than the original BERT and achieves 97% accuracy when tested on question-answering tasks. It maintains the original architecture of the BERT model while reducing the number of layers by a factor of two[16]. It is shown that DistilBERT achieves a similar performance for the named entity recognition of medical text with half the runtime and approximately half the disk space as the medical version of BERT[19]. ResNet-50 is a deep convolutional neural network with 50 layers. The ResNet-50 utilizes a skip connection between every two layers. The ResNet-50 model has shown superior performance in several medical image analysis[20–22] and also in dental image analysis[23,24]. Therefore, the DistilBERT and ResNet-50 models were utilized as encoders due to their extensive adaptation and irrefutable performance on medical data and image analysis.

**Loss Function:** First, the text encoder function converted the text into a text vector of size *n* and then a non-linear text projection function transformed the text vector into a vector of size *d* (Equation 1 and Equation 2) to generate text embedding. The *text projection function* is a fully connected neural network with *n* input and *d* output. Similarly, the image encoder function converts the image into an image vector of size *m* and then a non-linear image projection function converts it to the vector of size *d* (Equation 3 and Equation 4) to generate image embedding. Here, the *image projection function* is a fully connected neural network with *m* input and *d* output.

$$[P_{i1}\ P_{i2}\ P_{i3}\ ...\ P_{in}] = ITE(A_i) \quad [1]$$

$$Embedded\ Text\ (ET) = TP([P_{i1}\ P_{i2}\ P_{i3}\ ...\ P_{in}]) \quad [2]$$

Here, we use an initial text embedding function *ITE()* to generate the inputs to a text project function *TP()* to obtain the embedded text (ET).

$$[Q_{i1}\ Q_{i2}\ Q_{i3}\ ...\ Q_{in}] = IIE(B_i) \quad [3]$$

$$Embedded\ Image\ (EI) = IP([Q_{i1}\ Q_{i2}\ Q_{i3}\ ...\ Q_{in}]) \quad [4]$$

Similarly, we use an initial image embedding function *IIE()* to generate the inputs to a text project function *IP()* to obtain the embedded image (EI). Next, the cosine similarity function (Equation 5) was utilized to calculate the similarity between the embedded texts and embedded images (Equation 6, 7, and 8). The cosine similarity groups the data together based on their contents. It is better than Euclidian distance for image and text analysis[25]. Because sometimes two similar data (text or image) may far apart by Euclidian distance due to size but may closely be oriented for their contents. It finds the angle between two data. If the angles between two data are smaller, then the contents are similar and have higher cosine similarity.

$$cosine\ similarity(A, B) = \frac{A.B}{|A||B|} \ for\ two\ vectors\ A\ and\ B \quad [5]$$

$$Image\ Text\ Similarity\ (ETS) = cosine(ET, EI^T) \quad [6]$$

$$Text\ Similarity\ (TS) = cosine(ET, ET^T) \quad [7]$$

$$Image\ Similarity\ (IS) = cosine(EI, EI^T) \quad [8]$$

We then defined $targets = softmax(\frac{TS+IS}{2})$, where the $softmax$ function is:

$$softmax = \frac{e^{z_i}}{\sum_{j=1}^{K} e^{z_j}}; i = 1, \dots, K\ and\ z = (z_1, \dots, z_K) \in R^K\ \forall vector\ z \quad [9]$$

Finally, we calculated the $Total\ Loss = \frac{text\ loss\ +\ image\ loss}{2}$ using the text loss and image loss. The text and image losses are: $text\ loss = binary\ cross\ entropy\ loss\ (ETS, targets)$ and
$image\ loss = binary\ cross\ entropy\ loss = (ETS^T, targets^T)$

The $binary\ cross\ entropy\ loss$ is defined in Equation 10 where $y_i$ true class and $\hat{y}_i$ is the model's predicted class.

$$binary\ cross\ entropy\ loss = -\frac{1}{batch\ size} \sum_{i}^{batch\ size} y_i * log\hat{y}_i + (1 - y_i) * log(1 - \hat{y}_i) \quad [10]$$

The $Total\ Loss$ would be lower for the true image-text pairs and higher for the randomly coupled image-text pairs.

### 2.5 Data Preprocessing and Model Training

The dataset was randomly divided into 80% for training, 10% for validation, and 10% for model testing. The cohorts utilized for testing were different from the training and validation cohorts to avoid data snooping. Furthermore, the original and augmented image-text pairs resided in the same group, either training, validation, or testing to avoid any data leakage. The images were resized ($224 \times 224$) to reduce image heterogeneity, normalized (range 0-1) to remove variable range, and loaded using the RGB channels to preserve all information.

Pre-trained DistilBERT[16] and ResNet-50[17] models were utilized. The encoded texts were the vectors of size 768, encoded images were the vectors of size 2048, and after projection the final texts and images embeddings were the vectors of size 256. The maximum length of the texts/captions 200 characters, learning rate 0.0001, batch size 32, and epoch size 100 were utilized during model training.

### 2.6 Ablation Study

To understand the importance of paired text-image embedding for learning visual representation, we conducted an ablation study in two ways:

**Only Text Embedding Model:** We applied the pretrained DistilBERT[16] model for text embedding. First, we converted all the captions and queries into the vectors of size 768. Then, we computed the cosine similarity between the query vector and all the caption vectors and found the lowest distance. Finally, the image corresponding to the lowest value was extracted as the matched image. Figure 4 illustrates the image searching process using only text embedding models.

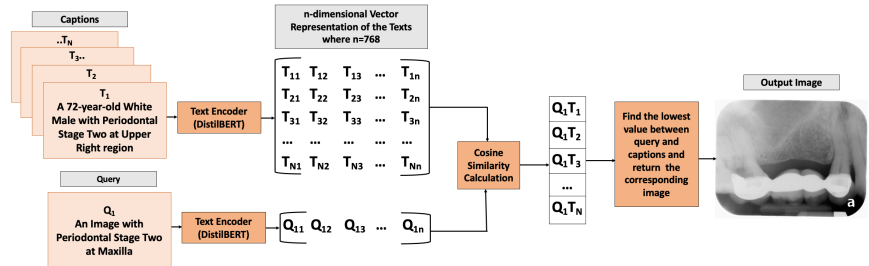

**Figure 4:** The workflow of the text embedding architecture.

**Only Image Embedding Model:** We applied the pretrained ResNet-50[17] model on periapical radiographs for image embedding. First, we converted all the images from the database and the query images into vectors of size 2048. Then, we computed the cosine similarity between the query image vector and all the database image vectors and found the lowest distance. Finally, the image corresponding to the lowest value was extracted as a matched image. Figure 5 illustrates the image searching process using only the image embedding framework.

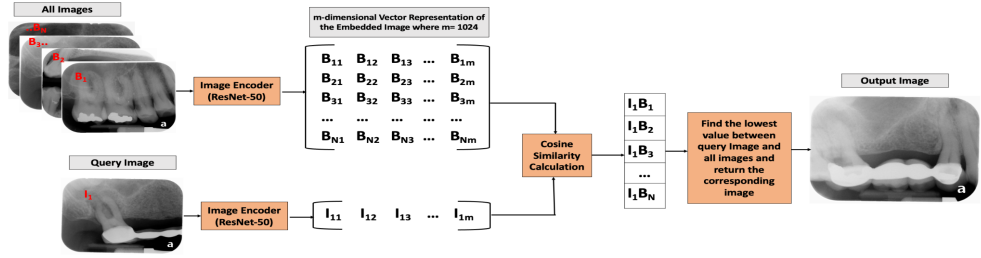

**Figure 5:** The workflow of the image embedding model.

## 2.7 Performance Evaluation Matrices

The model performance was evaluated using the hit@k (Equation 11), Precision@k (Equation 12), and Mean Reciprocal Rank (MRR) (Equation 13). Hit@k quantifies if we get any relevant result in the top-K returns. Precision@k quantifies how many items in the top-K results were relevant. MRR computes the rank of the first relevant item to a sample query. In addition, to provide the acceptability, reliability, and consistency of the proposed model, we have calculated Cohen's Kappa values between the annotators and the Dental CLAIRES model.

$$Hit@k = Any\ relevant\ image\ found\ on\ first\ k\ retrieved\ images \quad [11]$$

$$Precision@k = \frac{True\ Positive@k}{True\ Positive@k + False\ Positive@k} \quad [12]$$

$$Mean\ Reciprocal\ Rank\ (MRR) = \frac{1}{|Q|}\sum_{i=1}^{|Q|}\frac{1}{rank_i} \quad [13]$$

Here, $|Q|$ denotes the total numbers of queries and $rank_i$ denotes the rank of the first relevant result.

## 3. Results

### 3.1 Model Performance Evaluation

We calculated the hit and precision of the models on first, second, and third returns. We observed that the image and text augmentation process helped to improve the model performance, as illustrated in Table 2. We also speculated that contemplating the first three returns showed a more functional and robust model than only considering the first return. As it is shown in Table 2, our proposed model outperformed the other two models in all performance evaluation metrics.

**Table 2:** The performance comparison of the proposed model, text embedding model, and image embedding model.

| Model Name | Hit | | | Precision | | | MRR |
|---|---|---|---|---|---|---|---|
| | @1 | @2 | @3 | @1 | @2 | @3 | |
| **(A) Proposed Dental CLAIRES Model with Image and Text Augmentation** | **0.72** | **0.85** | **0.96** | **0.72** | **0.64** | **0.66** | **0.82** |
| (B) Dental CLAIRES Model without Image and Text Augmentation | 0.67 | 0.73 | 0.79 | 0.67 | 0.57 | 0.54 | 0.74 |
| (C) Only Text Embedding | 0.66 | 0.70 | 0.77 | 0.66 | 0.62 | 0.59 | 0.70 |
| (D) Only Image Embedding | 0.16 | 0.21 | 0.29 | 0.16 | 0.11 | 0.11 | 0.22 |

## 3.2 The Impact of Query Constraints on Model Performance

To understand the impact of text data constraints to find related images using our proposed model, we categorized queries based on the text data constraints-

- Low (only diagnosis)
- Medium (diagnosis with the region)
- Hard (diagnosis with the region and demographic information)

We evaluated 60 queries in each category. Table 3 demonstrates the hit, precision, and MRR of the proposed Dental CLAIRES model for different difficulty levels of the queries.

**Table 3:** Model's accuracy and precision for different difficulty levels.

| Difficulty Level | Examples | Hit | | | Precision | | | MRR |
|---|---|---|---|---|---|---|---|---|
| | | @1 | @2 | @3 | @1 | @2 | @3 | |
| Low | An image with Periodontal Stage Two. | 0.82 | 1.0 | 1.0 | 0.82 | 0.77 | 0.74 | 0.86 |
| Medium | An image with Periodontal Stage Two at the Left Lower Molar region. | 0.64 | 0.75 | 0.89 | 0.64 | 0.59 | 0.61 | 0.74 |
| Hard | A 49-year-old White Female with Periodontal Stage Two at Lower Molar Left region. | 0.70 | 0.90 | 1.0 | 0.70 | 0.68 | 0.41 | 0.86 |

Figure 6 illustrates a visual representation of the Dental CLAIRES model results on some queries with different difficulty levels and their corresponding output images.

**Difficulty Level Low: An image with Periodontal Stage Two.**

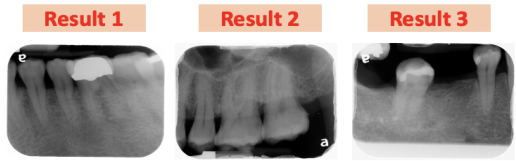

**Difficulty Level Medium: An image with Periodontal Stage one at Left Lower Molar region.**

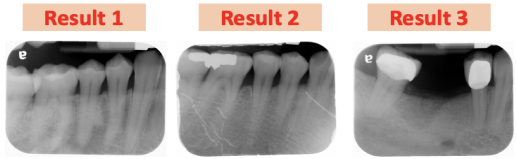

**Difficulty Level Hard: A white male with Periodontal Stage Three at Upper Anterior region.**

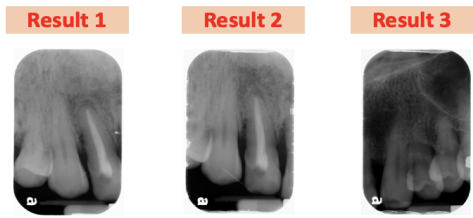

**Figure 6:** Dental CLAIRES' results on some queries for different difficulty levels.

## 3.3 Model's Consistency with Annotators

Table 4 demonstrates the agreement between the annotators and model's output. We utilized majority voting to get the periodontal stages of radiographs if there is a disagreement among annotators. We observed that the proposed model showed substantial agreement, $\kappa = 0.80$ with majority voting.

**Table 4:** Cohen's Kappa values between the annotators and the Dental CLAIRES model.

|  | Annotator 1 | Annotator 2 | Annotator 3 | Majority Voting | Model's output |
|---|---|---|---|---|---|
| Annotator 1 | 1 | 0.23 | 0.49 | 0.51 | 0.39 |
| Annotator 2 | 0.23 | 1 | 0.36 | 0.44 | 0.48 |
| Annotator 3 | 0.49 | 0.36 | 1 | 0.79 | 0.65 |
| Majority Voting | 0.51 | 0.44 | 0.79 | 1 | **0.80** |
| Model's output | 0.39 | 0.48 | 0.65 | **0.80** | 1 |

## 3.4 Graphical User Interface of the Proposed Model

Figure 7 illustrates the layout of the GUI of the proposed framework. The main window has three sections: "query", "number of images", and "output". The "query" section is used to load users' input, the "number of images" dropdown selection section allows users to decide how many images to return, and the "output" section will display retrieved images. Users can search for images of periodontal disease stages, regions, and demographic information based on their specific needs. The search engine is designed to be user-friendly, so that people can find the images they need quickly and easily.

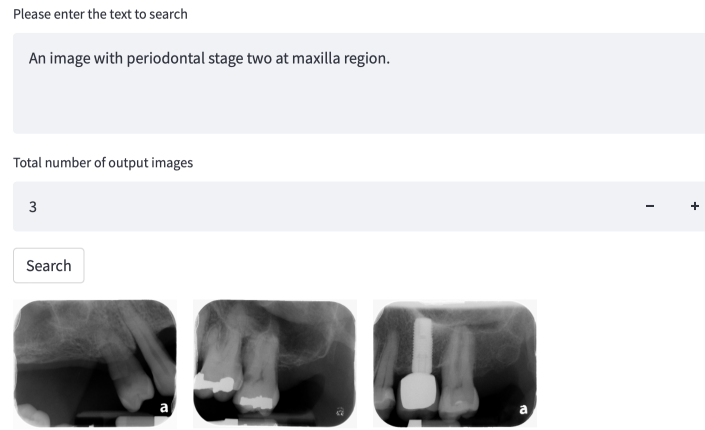

**Figure 7:** The graphical user interface (GUI) of the proposed Dental CLAIRES' model.

## 4. Discussion

The proposed search engine contains periodontal diagnosis and demographic information associated with periapical radiographs. We have utilized the image and text augmentation method to improve the model performance. The model achieved substantial agreement with the most experienced examiner (Annotator #3). Furthermore, the low Cohen Kappa coefficient ($\kappa = 0.23$) indicates that the deciding periodontal stages on periapical radiographs depend on the examiner's knowledge and experiences. So, there is an utmost need to develop an operator-independent tool that aids dentists in reliably assessing periodontitis, and our model can meet these needs.

We currently use a limited model for evaluating patients, which includes periodontal diagnosis and periapical radiographs. However, we hope to integrate other dental abnormalities, image modalities, and clinical data into our database in the future. This will include information such as the patient's medical history, vitals, and medications. In addition, we hope to be able to integrate data from other hospitals in order to create a more robust and generalizable model. This will allow us to incorporate images' visual features with the patient's clinical data.

Medical image is a significant data source in healthcare and an essential tool for diagnosing, monitoring, treatment planning, and understanding the underlying mechanism for disorders in the human body[26,27]. However, interpreting the medical image is challenging due to its complex nature and scarcity of appropriately annotated data[9]. Besides, human interpretation of medical images depends on interpreters' knowledge and experiences[28]. Furthermore, it is error-prone- one in four patients receive a false-positive diagnosis by image readings[28]. Therefore, it is necessary to generate high-quality annotated data for secondary research and train new trainees. However, generating high-quality data annotated by domain experts is expensive and time-consuming[9]. The proposed framework can be utilized as an automated annotation tool to assist human annotators and help to understand medical image data with systemic use of the textual data. Besides, clinicians and researchers can easily extract information depending on their goals and can be applied for patient care, research, and innovation using the user interface.

## 5. Conclusion

We introduced Dental CLAIRES (Contrastive LAnguage Image REtrieval Search for dental research), the first framework to retrieve radiographic images with periodontitis based on a user's query. The Cohen's Kappa score in Table 4 demonstrates that human experts have a large variability of agreements for periodontal staging, which shows the necessity of an automated system in dental research. We expect our work can motivate future work to make efficient and systemic use of textual data in imaging interpretation and understanding. It can also aid clinicians for clinical decision making and improved patient care.

**Acknowledgments**

TK is a CPRIT Predoctoral Fellow in the Biomedical Informatics, Genomics and Translational Cancer Research Training Program (BIG-TCR) funded by Cancer Prevention & Research Institute of Texas (CPRIT RP210045). XJ is CPRIT Scholar in Cancer Research (RR180012), and he was supported in part by Christopher Sarofim Family Professorship, UT Stars award, UTHealth startup, the National Institute of Health (NIH) under award number R01AG066749 and U01TR002062, and the National Science Foundation (NSF) #2124789.

**References**

1  iData Research. How Many Dental X-Rays are Performed in the United States? iData Research. 2019.https://idataresearch.com/how-many-dental-x-rays-are-performed-in-the-united-states/ (accessed 14 Jul2022).
2  Website. iData Research. How Many Dental X-Rays are Performed in the United States? iData Research https://idataresearch.com/how-many-dental-x-rays-are-performed-in-the-united-states/ (2019).
3  Valenza JA, Walji M. Creating a searchable digital dental radiograph repository for patient care, teaching and research using an online photo management and sharing application. *AMIA Annu Symp Proc* 2007; : 1143.
4  Walji MF, Spallek H, Kookal KK, Barrow J, Magnuson B, Tiwari T *et al.* BigMouth: development and maintenance of a successful dental data repository. *J Am Med Inform Assoc* 2022; **29**: 701–706.
5  Radford A, Kim JW, Hallacy C, Ramesh A, Goh G, Agarwal S *et al.* Learning Transferable Visual Models From Natural Language Supervision. 2021. doi:10.48550/arXiv.2103.00020.
6  Quattoni A, Collins M, Darrell T. Learning Visual Representations using Images with Captions. 2007 IEEE Conference on Computer Vision and Pattern Recognition. 2007. doi:10.1109/cvpr.2007.383173.
7  Joulin A, Grave E, Bojanowski P, Mikolov T. Bag of Tricks for Efficient Text Classification. Proceedings of the 15th Conference of the European Chapter of the Association for Computational Linguistics: Volume 2, Short Papers. 2017. doi:10.18653/v1/e17-2068.
8  Dong J, Li X, Snoek CGM. Predicting Visual Features From Text for Image and Video Caption Retrieval. IEEE Transactions on Multimedia. 2018; **20**: 3377–3388.
9  Zhang Y, Jiang H, Miura Y, Manning CD, Langlotz CP. Contrastive Learning of Medical Visual Representations from Paired Images and Text. 2020. doi:10.48550/arXiv.2010.00747.
10 Xu S, McCusker J, Krauthammer M. Yale Image Finder (YIF): a new search engine for retrieving biomedical images. *Bioinformatics* 2008; **24**: 1968–1970.
11 Eslami S, de Melo G, Meinel C. Does CLIP Benefit Visual Question Answering in the Medical Domain as Much

as it Does in the General Domain? arXiv [cs.CV]. 2021.http://arxiv.org/abs/2112.13906.
12  Mori M, Ariji Y, Fukuda M, Kitano T, Funakoshi T, Nishiyama W *et al.* Performance of deep learning technology for evaluation of positioning quality in periapical radiography of the maxillary canine. *Oral Radiol* 2022; **38**: 147–154.
13  Le-Khac PH, Healy G, Smeaton AF. Contrastive Representation Learning: A Framework and Review. IEEE Access. 2020; **8**: 193907–193934.
14  Schwendicke F, Singh T, Lee J-H, Gaudin R, Chaurasia A, Wiegand T *et al.* Artificial intelligence in dental research: Checklist for authors, reviewers, readers. *J Dent* 2021; **107**: 103610.
15  Tonetti MS, Greenwell H, Kornman KS. Staging and grading of periodontitis: Framework and proposal of a new classification and case definition. Journal of Periodontology. 2018; **89**: S159–S172.
16  Sanh V, Debut L, Chaumond J, Wolf T. DistilBERT, a distilled version of BERT: smaller, faster, cheaper and lighter. 2019. doi:10.48550/arXiv.1910.01108.
17  He K, Zhang X, Ren S, Sun J. Deep Residual Learning for Image Recognition. 2015. doi:10.48550/arXiv.1512.03385.
18  Devlin J, Chang M-W, Lee K, Toutanova K. BERT: Pre-training of Deep Bidirectional Transformers for Language Understanding. 2018. doi:10.48550/arXiv.1810.04805.
19  Abadeer M. Assessment of DistilBERT performance on Named Entity Recognition task for the detection of Protected Health Information and medical concepts. Proceedings of the 3rd Clinical Natural Language Processing Workshop. 2020. doi:10.18653/v1/2020.clinicalnlp-1.18.
20  Sarwinda D, Paradisa RH, Bustamam A, Anggia P. Deep Learning in Image Classification using Residual Network (ResNet) Variants for Detection of Colorectal Cancer. Procedia Computer Science. 2021; **179**: 423–431.
21  Guo S, Yang Z. Multi-Channel-ResNet: An integration framework towards skin lesion analysis. Informatics in Medicine Unlocked. 2018; **12**: 67–74.
22  Rajpal S, Lakhyani N, Singh AK, Kohli R, Kumar N. Using handpicked features in conjunction with ResNet-50 for improved detection of COVID-19 from chest X-ray images. *Chaos Solitons Fractals* 2021; **145**: 110749.
23  Mahdi FP, Motoki K, Kobashi S. Optimization technique combined with deep learning method for teeth recognition in dental panoramic radiographs. *Sci Rep* 2020; **10**: 19261.
24  Cejudo JE, Chaurasia A, Feldberg B, Krois J, Schwendicke F. Classification of Dental Radiographs Using Deep Learning. *J Clin Med Res* 2021; **10**. doi:10.3390/jcm10071496.
25  Sohangir S, Wang D. Improved sqrt-cosine similarity measurement. *J Big Data* 2017; **4**. doi:10.1186/s40537-017-0083-6.
26  Machine learning in translational bioinformatics. In: *Translational Bioinformatics in Healthcare and Medicine*. Academic Press, 2021, pp 183–192.
27  Havaei M, Guizard N, Larochelle H, Jodoin P-M. Deep Learning Trends for Focal Brain Pathology Segmentation in MRI. Lecture Notes in Computer Science. 2016; : 125–148.
28  Anbil PS, Ricci MT. Consider the Promises and Challenges of Medical Image Analyses Using Machine Learning. mddionline.com. 2020.https://www.mddionline.com/radiological/consider-promises-and-challenges-medical-image-analyses-using-machine-learning (accessed 11 Jul2022).